\documentclass[conference]{IEEEtran}
\IEEEoverridecommandlockouts
\usepackage{cite}
\usepackage{amsmath,amssymb,amsfonts}
\usepackage{graphicx}
\usepackage{textcomp}
\usepackage{xcolor}
\usepackage{amsmath}
\usepackage{algorithm}
\usepackage{algorithmicx}
\usepackage{algpseudocode}
\usepackage{amsthm,amsmath,amssymb}
\usepackage{graphicx}
\usepackage{subfigure}
\usepackage{booktabs}
 \usepackage[switch]{lineno}
\def\BibTeX{{\rm B\kern-.05em{\sc i\kern-.025em b}\kern-.08em
    T\kern-.1667em\lower.7ex\hbox{E}\kern-.125emX}}
\begin{document}

\title{Adaptive Adversarial Training for Meta Reinforcement Learning\\
}
\author{
\IEEEauthorblockN{Shiqi Chen${^{1,3}}$, Zhengyu Chen${^{1,2}}$, Donglin Wang${^{1,2}}$\IEEEauthorrefmark{1}} 
  
\IEEEauthorblockA{${^{1}}$Machine Intelligence Lab (MiLAB), AI Division, School of Engineering, Westlake University} 
\IEEEauthorblockA{${^{2}}$Institute of Advanced Technology, Westlake Institute for Advanced Study} 
\IEEEauthorblockA{${^{3}}$Wee Kim Wee School of Communication and Information, Nanyang Technological University } 

\thanks{
 \textsuperscript{$\ast$}Corresponding author.}
}

\maketitle

\begin{abstract}
Meta Reinforcement Learning (MRL) enables an agent to learn from a limited number of past trajectories and extrapolate to a new task.
In this paper, we attempt to improve the robustness of MRL. We build upon model-agnostic meta-learning (MAML) and propose a novel method to generate adversarial samples for MRL by using Generative Adversarial Network (GAN). That allows us to enhance the robustness of MRL to adversal attacks by leveraging these attacks during meta training process.
\end{abstract}

\begin{IEEEkeywords}
Adversarial Training, Meta Reinforcement Learning, GAN, Robustness.
\end{IEEEkeywords}

\section{Introduction}

\noindent A major concern in machine learning is the generalization from limited samples, and meta learning is a candidate solution to extract knowledge from a limited number of data and solve unseen problems quickly. In the past research of reinforcement learning, Meta Reinforcement Learning (MRL) \cite{b4} has shown strong advantage in extracting information from a limited number of trajectories, and related research has been advancing greatly due to rapid progress on both deep reinforcement learning and meta learning algorithms \cite{b11, b20}.
For example, Finn et al.\cite{b4} propose a Model Agnostic Meta Learning (MAML) framework that learns from past trajectories by using policy gradient in the context of reinforcement learning. They train an agent with a set of initial parameter that has gratifying generalization ability, i.e. when this agent approaches new tasks, it only requires a few gradient steps to address these new tasks. This method is model-agnostic because it can potentially be applied to arbitrary algorithms that optimize by using gradient descent.

Robustness is another major concern for machine learning. While MRL shows great progress in recent years, few researchers pay attention on the robustness problem of MRL when it faces attacks. However, MRL, just like other RL methods in general, is quite vulnerable to a few different problems. First, it is vulnerable to adversarial manipulation of training data (states or rewards in MDP process), which is an inherent problem of deep neural networks\cite{b13}\cite{b28}. Second, in the meta learning scenario, the task data set is usually small, and the agent experiences only a limited number of trajectories.Therefore, MRL is easily affected by noises and has large errors when confronting adversarial attacks. In practice, plenty of former research indicates that the noise, including both the naive attacks and adversarial attacks, can severely degrade the performance of reinforcement learning algorithm. For example, perturbations are quite common in robotics locomotion, such as wind and other natural forces. These perturbations are channeled into images and are subsequently magnified by RL. This paper attempts to improve the robustness of the algorithm and address these corruptions.

The primary contribution of our work is to learn a more robust initial parameter to resist adversarial attack for MRL. And our contributions are as follows:

\begin{itemize}
\item First, we represent a novel objective function of adversarial attack specifically in reinforcement learning setting whose optimization degrades the performance of RL algorithms. Then we apply our approach on the meta reinforcement learning scenario, our experiment shows our approach could significantly reduce the total average return of MRL to a lower rank, it also indicates that the attack fools an agent to take the possible worse actions in its current state in greater possibility.
\item To the best knowledge of the authors, this is the first work to resolve the robust problem under MRL setting. We aim to find a solution to help the agent resist adversarial attack more efficiently. We evaluate our approach in five different environments:  2D-navigation, Half-cheetah-Dir and Vel, Ant Dir and Vel. In our experiments, we show that our approach has better performance than MAML, naive training and the adversarial training by FGSM, which demonstrates that our method can improve the robustness of MRL system efficiently.
\item We propose a novel Generative Adversarial Network based approach to generate adversarial samples for current states in trajectories in reinforcement learning, which is shown as an efficient adversarial attack approach in our experiment. And what's more, this DNN based method has good adaptive ability that could be effective in different cases.
\item We propose a self-adaptive method to learn the "best" perturbation and the optimal policy simultaneously and we build an end-to-end framework to accomplish the goal of optimizing two objective functions simultaneously.
\end{itemize}

\section{Related Work}
\subsection{Reinforcement Learning}

Reinforcement learning, the reinforcement learning task $T_{i}$ is a Markov decision process\cite{b14}, is defined by a tuple $\left ( X, A, p, p_{0}, r, H \right )$. X indicates the state space, A the action space, $p(x_{t+1}\mid x_{t},a_{t})$ is the transition probability, $p_{0}$  is the initial state distribution. R is the reward function. And the on-policy method aims to learn a $f_{\theta }$ maping states $x_{t}$ to actions $a_{t}$. The goal of reinforcement learning is to learn a function $\pi$ parameterized by $\theta$ in such a way that it maximizes the expected total discounted reward \cite{b18}.
 The loss function is as follows:
\begin{equation}
L_{T_{i}}(f_{\theta })=-E_{x_{t},a_{t}\sim f_{\theta},q_{T_{i}}}\left[\sum_{t=1}^{H}R_{i}(x_{t},a_{t})\right]
\end{equation}

In the on-policy setting, such as REINFORCE\cite{b21} and TRPO\cite{b17}, we use the likelihood ratio policy gradient of the RL objective,
\begin{equation}
\begin{aligned}
\nabla_{\theta} J(\theta)&=\mathbb{E}_{\rho^{\pi}, \pi}\left[\nabla_{\theta} \log \pi_{\theta}\left(a_{t} \mid x_{t}\right)\left(\hat{Q}\left(x_{t}, a_{t}\right)-b\left(x_{t}\right)\right)\right]\\&=\mathbb{E}_{\rho^{\pi}, \pi}\left[\nabla_{\theta} \log \pi_{\theta}\left(a_{t} \mid x_{t}\right) \hat{A}\left(x_{t}, a_{t}\right)\right]
\end{aligned}
\end{equation}
Here, $\hat{Q}\left(x_{t},a_{t}\right)=\sum_{t^{\prime}=t}^{\infty} \gamma^{t^{\prime}-t} r\left(s_{t^{\prime}}, a_{t^{\prime}}\right)$ is the Monte Carlo estimate of the $Q^{\pi}\left(x_{t}, a_{t}\right)=\mathbb{E}_{s_{t+1}, a_{t+1}, \cdots \sim \pi \mid x_{t}, a_{t}}\left[\hat{Q}\left(x_{t}, a_{t}\right)\right]$, and $\rho^{\pi}=\sum_{t=0}^{\infty} \gamma^{t} p\left(x_{t}=s\right)$ are the innormalized state visitation frequencies, and the $b(x_{t})$ is the baseline to reduce the variance of the gradient estimate, in some cases we use the value function $V^{\pi}\left(x_{t}\right)=\mathbb{E}_{a_{t} \sim \pi\left(\cdot \mid x_{t}\right)}\left[Q^{\pi}\left(x_{t}, a_{t}\right)\right]$ to estimate the baseline $b(x_{t})$ , then $\hat{A}\left(x_{t}\right)$ is the estimate of advantage function $A\left(x_{t}\right)=Q^{\pi}\left(x_{t}, a_{t}\right)-V^{\pi}\left(x_{t}\right)$ which is the estimate of the advantage function.

\subsection{Meta Reinforcement Learning}

The combination of deep neural networks and reinforcement learning methods leads to huge advances for the agents to learn accurate policies for complex tasks. In conventional reinforcement learning, each agent learns a private policy in each task by interacting with the environment plenty of times. While when dealing with a large number of tasks, if each agent executes its training process separately, it would be slow and inefficiently. Learning large repertoires of goals by conventional reinforcement learning is prohibitive. In real cases, the tasks often share lots of common features, for instance going upstairs and downstairs both involve bending the knee and lifting the legs. Meta learning methods could exploit these common structure by collecting previous tasks and implement parameter generalization on meta-training set\cite{b26}\cite{b27}. If we hope the agents could quickly adapt to a new task, the meta reinforcement learning algorithm could help alleviate this problem by aggregating prior trajectories \cite{b3}\cite{b11}\cite{b20}.

Meta-learning aims to automatically learn the learning algorithms with satisfactory performance effectively. During the training process, these algorithms leverage data from previous tasks to acquire a learning method that could adapt to new tasks quickly. These methods have an assumption that the previous tasks and the new tasks are from the same distribution $\rho (T)$. In the supervised learning scenario, we hope to learn a function  $f_{\theta }$ with parameter $\theta$ that minimizes a supervised loss $L_{T}$. Then, our goal is to find a learning procedure, indicated as $\theta ^{'}=u_{\varphi }(D_{T}^{tr})$.
And thus, the overall loss function of meta-learning is as follows:
\begin{equation}
\begin{aligned}
\underset{\theta,\varphi }{min}E_{T\sim \rho (T)}\left [ L(D_{T}^{test},\theta ^{'}) \right ]\\
s.t.\quad
\theta ^{'}=u_{\varphi}(D_{T}^{tr},\theta)
\end{aligned}
\end{equation}
Among these, our method is based on model-agnostic meta learning \cite{b4}, which aims to learn a efficient initial parameter that can adapt to new task after one or several times gradient descent, MAML uses gradient descent as a learning algorithm:
\begin{equation}
u_{\varphi}(D_{T}^{tr},\theta)=\theta-\alpha\bigtriangledown _{\theta }L(D_{T}^{tr},\theta)
\end{equation}

These meta learning approaches also have wide application ground in the reinforcement learning scenario. In reality, the complex and diverse environments may have higher requirement for the generalization capacity of the agent. Recently, meta reinforcement learning has been developed for meta-learning dynamics models\cite{b12}\cite{b16}\cite{b2}and policies\cite{b4}\cite{b3} \cite{b11}where supervised meta reinforcement learning uses meta learning methods to learn a optimal policy which could adapt to new tasks quickly. In practice, the meta reinforcement learning process is a bit different from the supervised learning framework. Each task in MRL scenario is a different trajectory with the same action/state space but a different reward function.\cite{b4}
Our work is based on the gradient-based meta-learning framework under the reinforcement learning scenario, which learns from aggregated experience by policy gradients\cite{b1}\cite{b2}.

\subsection{Adversarial Training}

Adversarial examples are small perturbations of the original inputs, often imperceptible to a human observer, which could easily misguide the neural network into producing incorrect results.
Plenty of attack strategies to generate adversarial examples have been proposed in prior research. One of the most popular ways to engineer adversarial attacks on neural networks is the fast gradient sign method proposed by\cite{b5}, which crafts a cost function aiming at leading to decrease the network's performance. It takes into account a linear approximation of deep learning model and generates an attack. In practice, for instance a linear model,$f(x)=w^{T}x$, $x$ represents the input, $w$ represents the model parameter and $f(x)$ represents the output. We turn the normal sample $x$ into adversarial sample
\begin{equation}
x_{a}=x+\eta
\end{equation}
with $L_{\infty }$ constraints where
\begin{equation}
\eta=\alpha sign(\bigtriangledown _{x}l_{f}(x,y))
\end{equation}
Carlini and Wagner\cite{b1} then propose optimization on the perturbation by setting constrains. The objective function they try to minimize is:$ \left \| \lambda \right \|+\lambda l_{f}(x_{a},y)$.Where $\left \| \lambda  \right \|$ is a norm function, while it suffers from the efficiency problem and could only perturb one sample each time.

The adversarial attacks have been extended to deep reinforcement learning\cite{b6}\cite{b7}\cite{b9},Huang et al.\cite{b6} first propose a FGSM based framework to generate adversarial attacks on policies in the DRL. Kos et al.\cite{b7} present a improvement on Huang's method for reducing the number of attacks based on the value function, and only disrupting the agent in crucial moments when it is close to achieving a reward. Amin et al. \cite{b24} study a security threat to reinforcement learning where an attacker poisons the learning environment.

There are also some research concerning about the robust problem in meta learning in recent years. For instance, Lee et al. \cite{b8} proposed a meta-learning framework call Meta-Dropout which could learn to perturb the latent features of training examples for generalization. Nonetheless, the noise generator in this paper could only generates random noises which has weaker strength compared with the adversarial attack. Yin et al.\cite{b22} represents ADML to introduce adversarial samples to meta-learning, while it ignores the generating process of the adversarial samples. Micah et al.\cite{b25} adapt adversarial training for meta-learning and adapt robust architectural features to small networks for meta- learning.

\section{Preliminaries}
In the usual case, each task $T_{i}$ in the reinforcement learning is described as a Markov decision process with the horizon $H$,which contains an initial distribution $q_{i}(x_{1})$ and a transition distribution $q_{i}(x_{t+1}\mid x_{t},a_{t})$, the loss function $L_{T_{i}}$ corresponds to the (negative) reward function $R$. While in meta reinforcement learning scenario, the meta learner will sample multiple trajectories for few-shot learning, and the tasks belong to $p(T)$, the policy $f_{\theta }$ maps from the states $x_{t}$ to a distribution over actions $a_{t}$ at each time step $t\in \left \{ 1,2...H \right \}$. The loss for task $T_{i}$ is described as
\begin{equation}
 L_{T_{i}}(f_{\theta })=-E_{x_{t},a_{t}\sim f_{\theta },q_{T_{i}}}\left [ \sum_{t=1}^{H}R_{i}(x_{t},a_{t}) \right ]
\end{equation}
In our adversarial training scenario, our adversarial GAN could generate adversarial samples and perturbs normal states $ x_{t}$ to $\tilde{x_{t}}$. And the GAN parameter is indicated as $\phi$.
In $K$-shot reinforcement learning, we sample $K$ trajectories $(x_{1},a_{1}...x_{H})$ from $f_{\theta}$, and the corresponding rewards $R(x_{t},a_{t}$ for adaptation, then we use policy gradient methods to update the model gradient and optimize the meta-parameter.

\section{Methodology}

In this section, we present our proposed adMRL in detail. It includes two phases:  generating adversarial attack using adGAN and further leveraging them to improve the robustness of the MRL algorithm. The whole algorithm is is summarized in Algorithm 1. 

Figure 1 shows our overall framework of our model. Basi- cally we add an adGAN to MAML. In the meta training process, we use adGAN to generate noise to states both at training and validation process. And in the meta testing process, we test and evaluate our model with different noise types including random noise, FGSM noise and our adGAN noise.

\begin{figure}[h]
\centering
\includegraphics[width=0.5\textwidth]{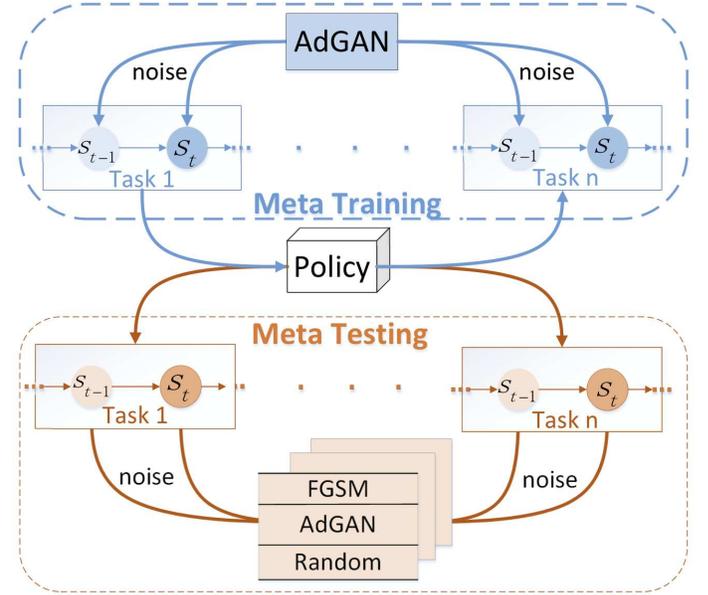} 
\caption{Interacting process during training, the adGAN generates the adversarial samples and misleads the agent to degrade the total rewards. By adversarial training, the policy has strong robustness. Then in the meta-test phase, the robust policy could give better performance under several attacks. }
\label{fig:1}
\end{figure}

\begin{algorithm}[htb]
    \caption{adMRL}
    \label{alg2}
	\begin{algorithmic}[1]
		\State Require: $p(T)=$ distribution over tasts. $\alpha,\beta$: step size hyperporameters.
        \State random initialize network parameter $\theta$, GAN parameter $\phi$.
		\While {not done do}
		\State Sample batch of tasks $T_{i}\sim p(T)$
        \For{all $T_{i}$}
        \State Using the current status and  generator of GAN to generate the polluted states.
        \State Sample K trajectories $D=\left \{ \left ( \tilde{x_{1}},a_{1},\cdot \cdot \cdot , \tilde{x_{H}}\right ) \right \}$ using $ f_{\theta }$  and $\tilde{\theta }$ in $T_{i}$. The perturbed states $\tilde{x_{i}}$ is generated by the generator of our adGAN, and the input are the normal states.
        \State Evaluate $\triangledown _{\theta}L_{T_{i}}(f_{\theta })$ using D and $ L_{T_{i}}$ in Eq.(1).
        \State Compute adapted parameters with gradient descent use Eq.(3).
        \State Sample K trajectories ${D}'=\left \{ \left ( {\tilde{x_{1}}}',{a_{1}}',\cdot \cdot \cdot , {\tilde{x_{H}}}'\right ) \right \}$ using $ f_{{\theta }'}$  in $T_{i}$. Meanwhile, the  perturbed states $\tilde{x_{i}}$ is generated by the generator of our adGAN, and the input is the normal states.
        \EndFor
        \State Update $\theta \leftarrow \theta-\beta \bigtriangledown _{\theta }\sum _{T_{i}\sim p(T) }L_{T_{i}}(f_{\theta })$ using each ${D_{i}}'$ and $L_{T^{i}}$
        \State Update $\phi$ using Eq.(5).
        \EndWhile

	\end{algorithmic}
\end{algorithm}

\subsection{adGAN for reinforcement learning}

In this subsection, we introduce the adGAN, which aims to generate adversarial attack that will degrade the performance of RL significantly. The key idea behind adGAN is corrupting each current states in RL, alluring the agent to believe it is in a correct state but will instead cause the agent to take a possible sub-optimal choice in each step and further decrease the total rewards.

\textbf{Definition 1} \emph{An adversarial attack for reinforcement learning is any possible perturbation that would possibly fool an agent to take "bad" actions which could finally cause lower total rewards}.

This definition is valid both for value function based algorithms and policy gradient based algorithms. In our experiment setting, MAML uses on-policy algorithms REINFORCE and TRPO. There is an important difference compared to supervised learning tasks such as image classification where attack is seen as successful if a given image is classified into a wrong label, while in a reinforcement learning setting, there is no concept of worst possible image but the agents can have worst average returns. From this point of view, we can design the loss function of adversarial attack under the reinforcement learning scenario as follows.
\begin{equation}
\begin{aligned}
\tilde{x}=\max _{x} L_{\text {Ti }}(f \varphi),\\
\|\tilde{x}-\bar{x}\| \leq \epsilon.
\end{aligned}
\end{equation}

Here, $\bar{x}$ represents the current optimal state in the Markov Decision Process, $\tilde{x}$ is the perturbed state, and $\epsilon$ is the constraint of perturbation range. To maximize the reinforcement learning function, each state is more possible to be corrupted into a bad action which may further cause an inappropriate action and finally get terrible performance.
First, we propose a method of generating adversarial attack in reinforcement learning, inspired by \cite{b23}, we can generate adversarial samples by an adGAN network, which consists of three parts: a generator $G$, a discriminator $D$, and the reinforcement learning model $F$. Here the generator takes the original states $x$ in the Markov decision process as its input and generates the perturbed state $ \tilde{x}$. Then all the perturbed state will be sent to the discriminator $D$, which could distinguish the "contaminated" states and the real states. To accomplish the goal of generating adversarial samples, we use the agents perform in our MRL model and obtain the  $L_{adv^{f}}$, which aims to mislead the agent into gaining worse returns than before.

The adversarial loss \cite{b5} can be written as follows:
\begin{equation}
L_{GAN}=E_{x}log D(x))+E_{x}log (1-D(x+G(x))).
\end{equation}
The discriminator D aims to distinguish the real samples $x$ and the "fake" samples generated by generator $G$. This is the loss function of a normal GAN which aims to generate samples in the same distribution with the original samples. The whole network achieves Nash equilibrium finally after the counterbalance process of generator and discriminator.

The loss for misleading the agent is:
\begin{equation}
L_{adv^{f}}=-L_{T_{i}}(f_{{\theta}'})(T_{i}\in {D_{i}}').
\end{equation}
Different from the advLoss proposed by \cite{b23}, we use the negative number of the loss function of reinforcement learning, in this way, the advLoss encourages the generator to generate the samples degrading the performance of agents. In specific, this loss could fool the agents to take the incorrect action in its current state and finally reduce the total returns.

The third loss is:
\begin{equation}
L_{hinge}=E_{x} max(0,\left \| G(x)_{2}-c \right \|).
\end{equation}
Where $c$ denotes the bound.
This loss is to bound the variation range of perturbation. In this case it avoids the samples differing too much with the original samples. Here, the c is a hyper-parameter to define the variation range.
Finally, our full object function is:
\begin{equation}
L=L_{adv^{f}} +\alpha L_{GAN}+\beta L_{hinge}.
\end{equation}
Here, the $\alpha$ and $\beta$ are hyper-parameters to control the relative importance of each sub objective function. By optimizing this loss function, we can get the adversarial samples which are from the same distribution as the "good" samples (controlled by $\alpha L_{GAN}$), but could arise bad performance (by optimizing $L_{adv^{f}}$), and the perturbation scale is constrained by $L_{hinge}$.

\subsection{Perturbation}
Our method is built on the basis of model agnostic meta learning \cite{b4}. Unlike the supervised learning setting where the support and query set are given by frigid dataset, during the MRL meta-training process with MAML-TRPO setting, we should sample the trajectories two times to serve as the role of support and query sets. The first time is to gain the loss for calculating the one-step policy gradient which plays the role of support set. And the second time uses the adapted policy to sample trajectories and gains the outer loss to update the initial policy parameters where they are similar to the query set and we call this process the validation phase. The meta-test maintains this procedure as well.

For adversarial attack, we inject the perturbation states only in the meta test process. Specifically, we use each current state as input to the adGAN and generate corresponding corruped states. Then, we use the polluted states to obtain the action for next time step.
For adversarial training, under this setting, we inject our adversarial samples in the meta-training process (including training and validation phases). Distinguished with MAML, we use our adGAN to generate  "false" observations which are the adversarial examples for each current states. In this case, the agent is able to learn initial parameters which have better generalization capacity due to accumulating knowledge in a more difficult environment. Thus, the agent could learn the ability to fight against these "bad" samples when facing an unseen new task.

\subsection{Optimization}
During the meta-training process, we have injected adversarial samples in both training and validation phases, where all these samples are used for the GAN loss and the Hinge loss backpropagation to distinguish the data from the true distribution and the fake data. For the adv loss, under the complexity consideration, we only use the outer loop loss to calculate the adv loss.

Thus, in our framework, there are in total two sets of parameters needed to be optimized: adGAN and the reinforcement learning model parameters. We treat both two sets of parameters as meta parameters and update them together. In this case, our adversarial samples also have good generalization capacity that could play a good role in test phase.

\section{Experiments}

\subsection{Simulation Environments}

\begin{figure*}[t]
 \centering
 \subfigure[2D-Navigation]{
  \begin{minipage}[t]{0.19\linewidth}
   \centering
   \includegraphics[scale=0.17]{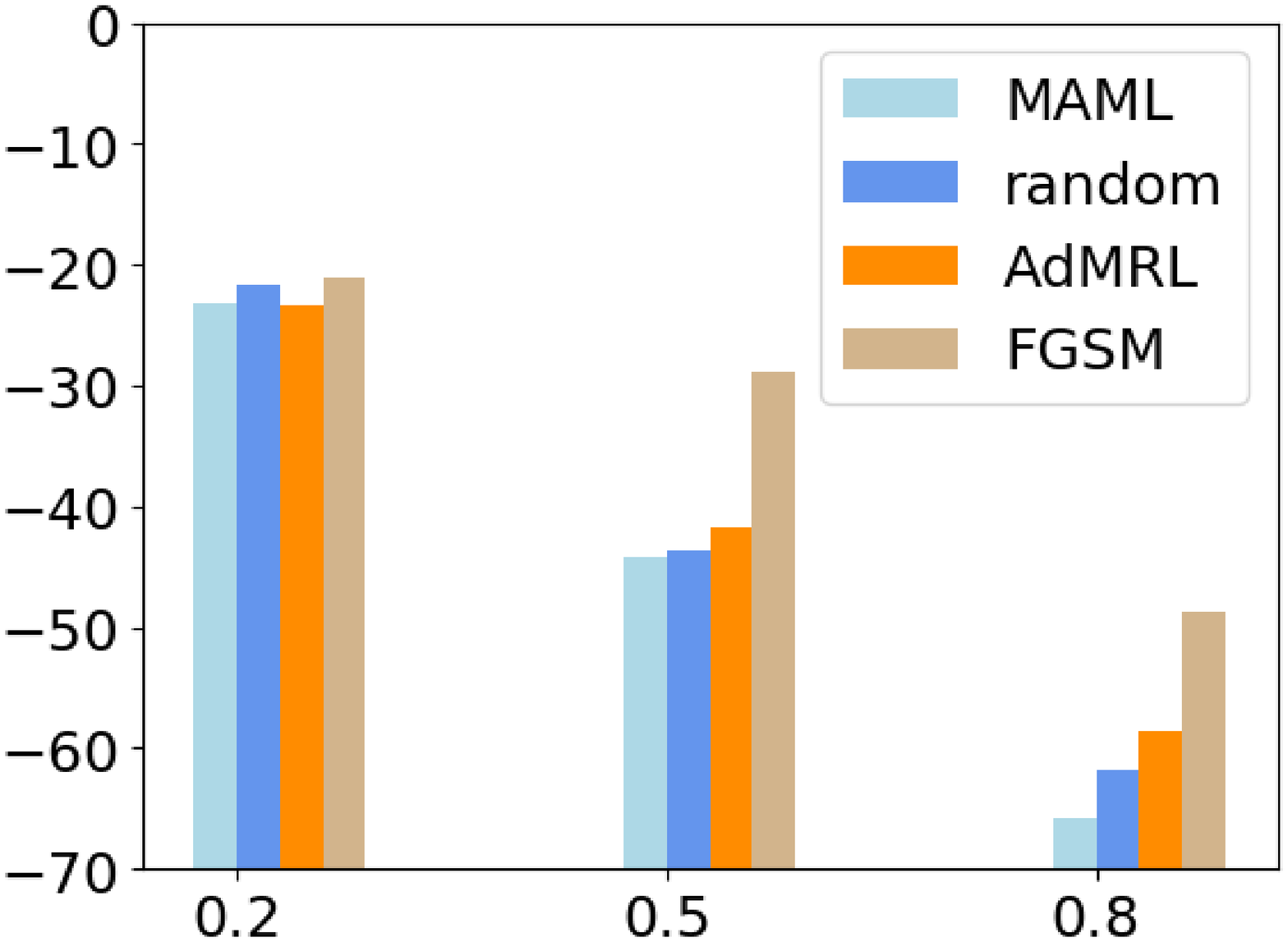}
  \end{minipage}%
 }%
 \subfigure[HalfCheetah-Vel]{
  \begin{minipage}[t]{0.19\linewidth}
   \centering
   \includegraphics[scale=0.17]{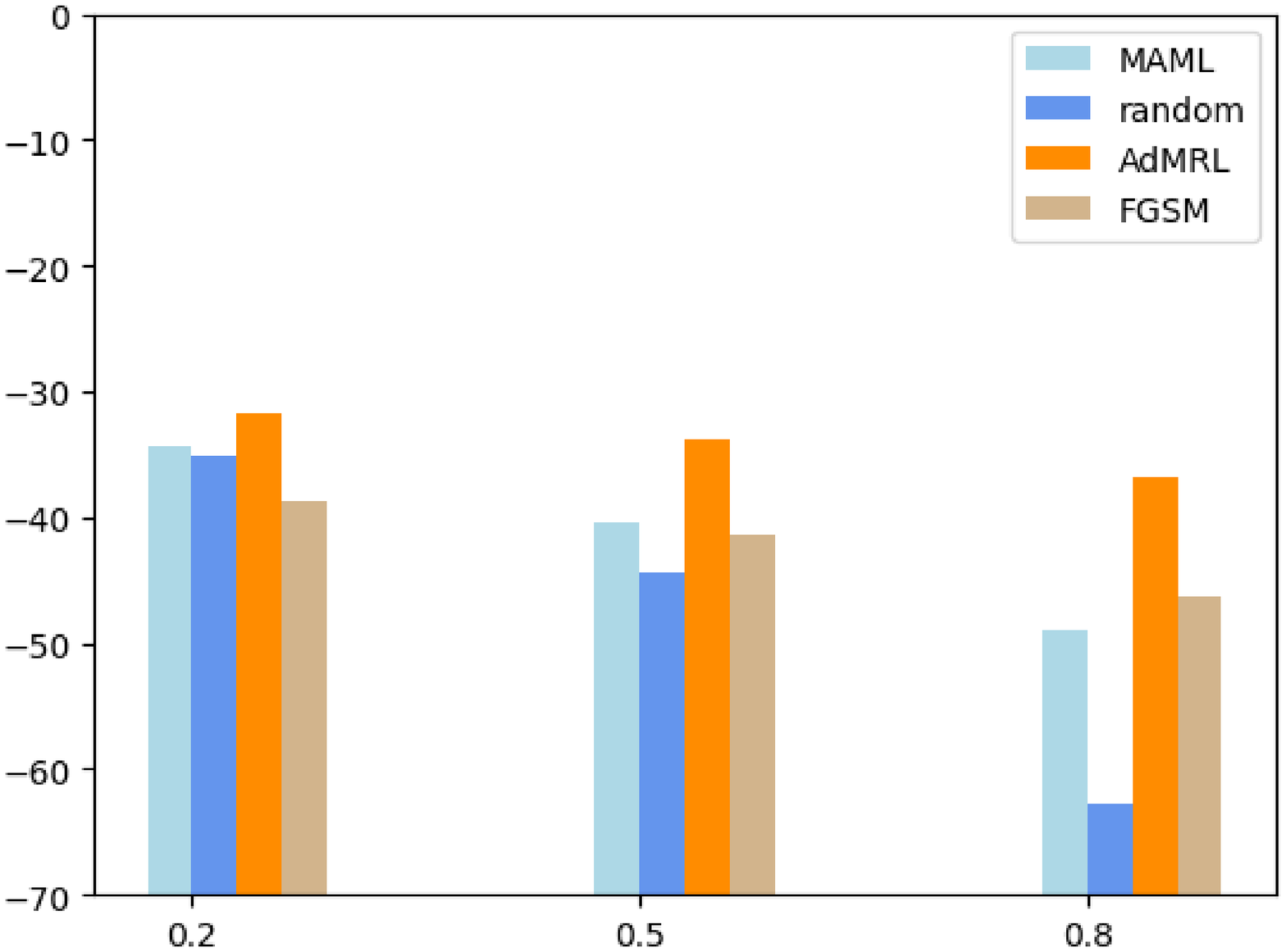}
  \end{minipage}%
 }%
 \subfigure[HalfCheetah-Dir]{
  \begin{minipage}[t]{0.19\linewidth}
   \centering
   \includegraphics[scale=0.17]{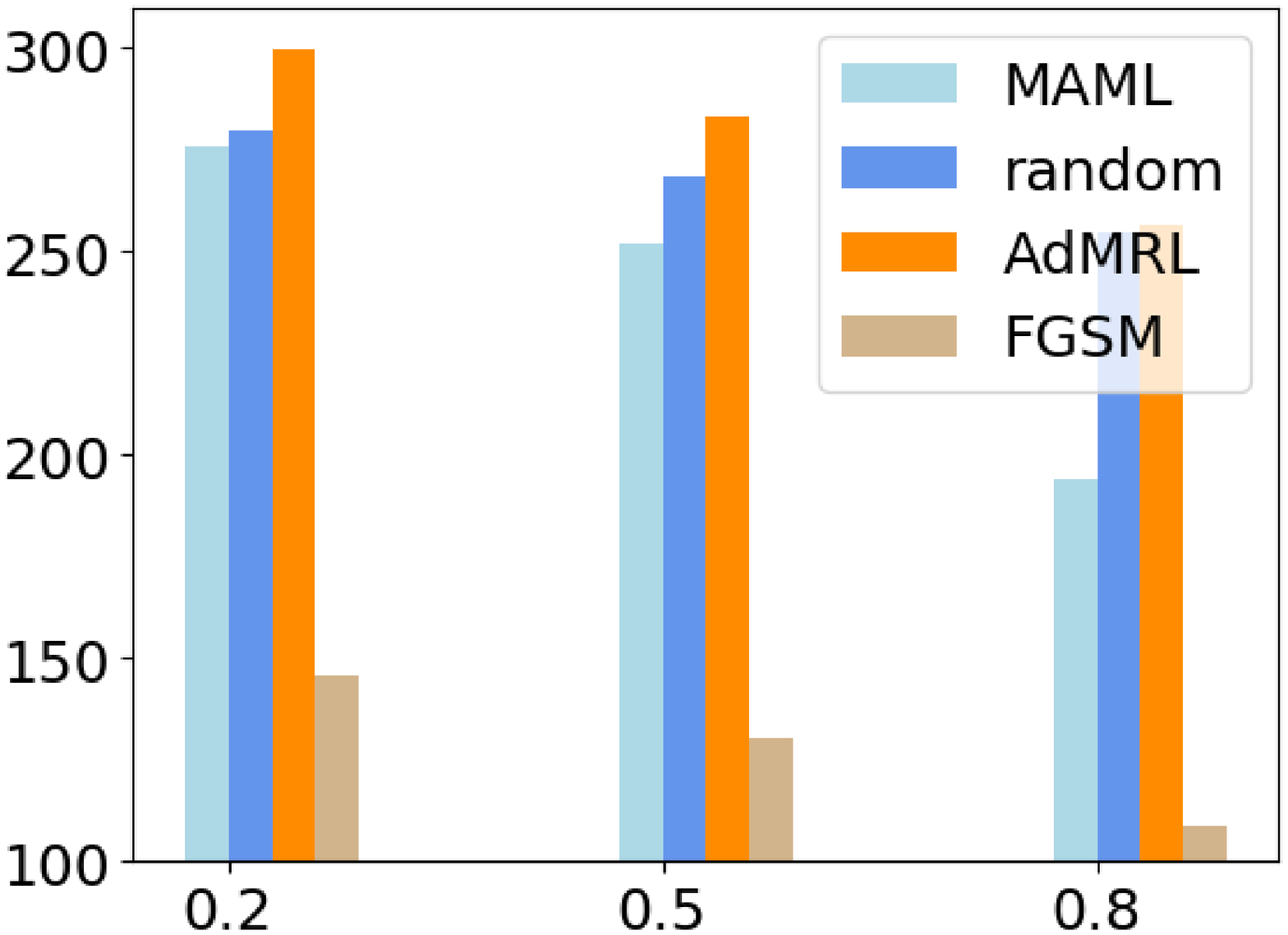}
  \end{minipage}
 }%
 \subfigure[Reacher]{
  \begin{minipage}[t]{0.19\linewidth}
   \centering
   \includegraphics[scale=0.17]{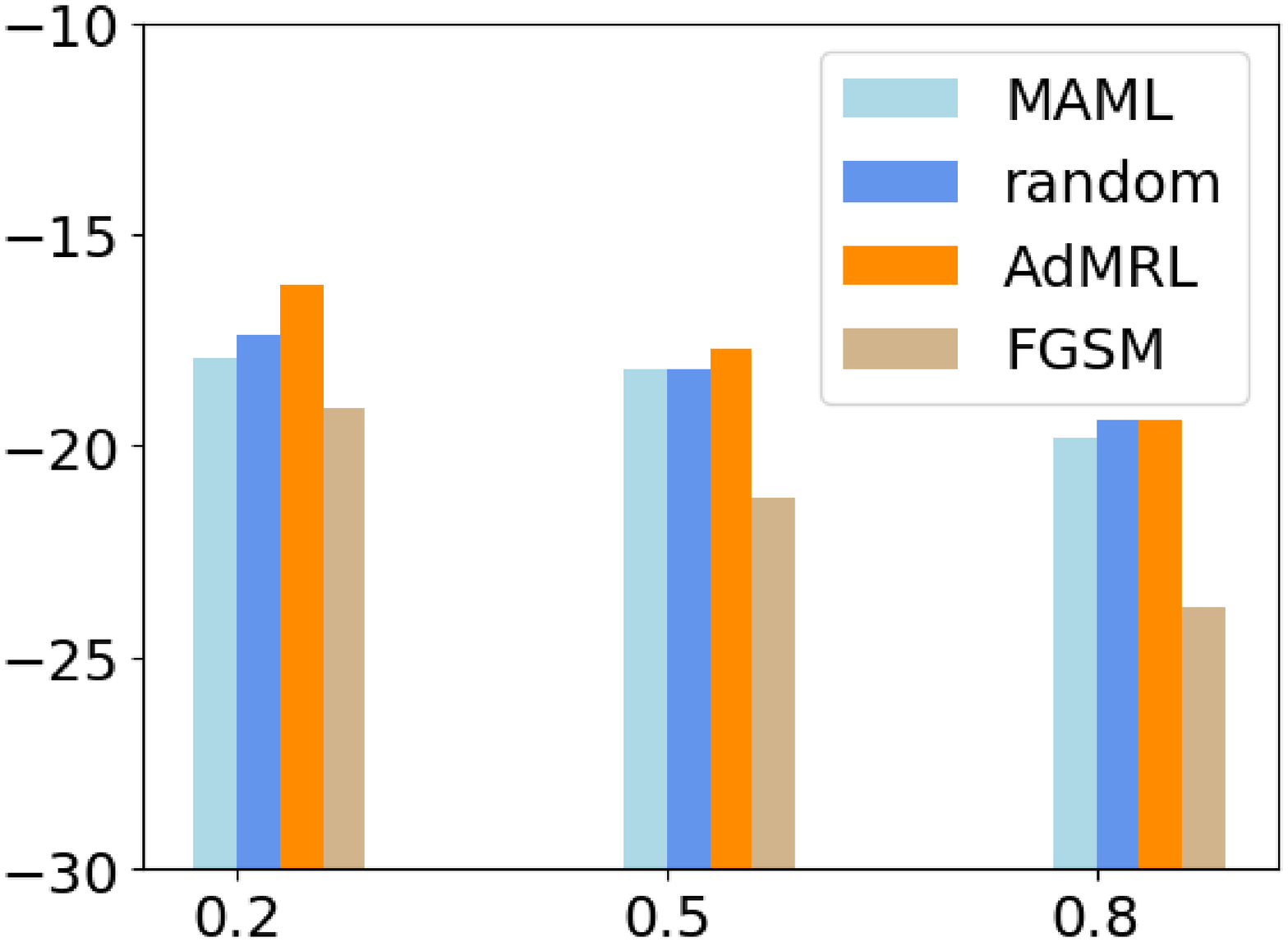}
  \end{minipage}
 }%
 \subfigure[Ant]{
  \begin{minipage}[t]{0.19\linewidth}
   \centering
   \includegraphics[scale=0.17]{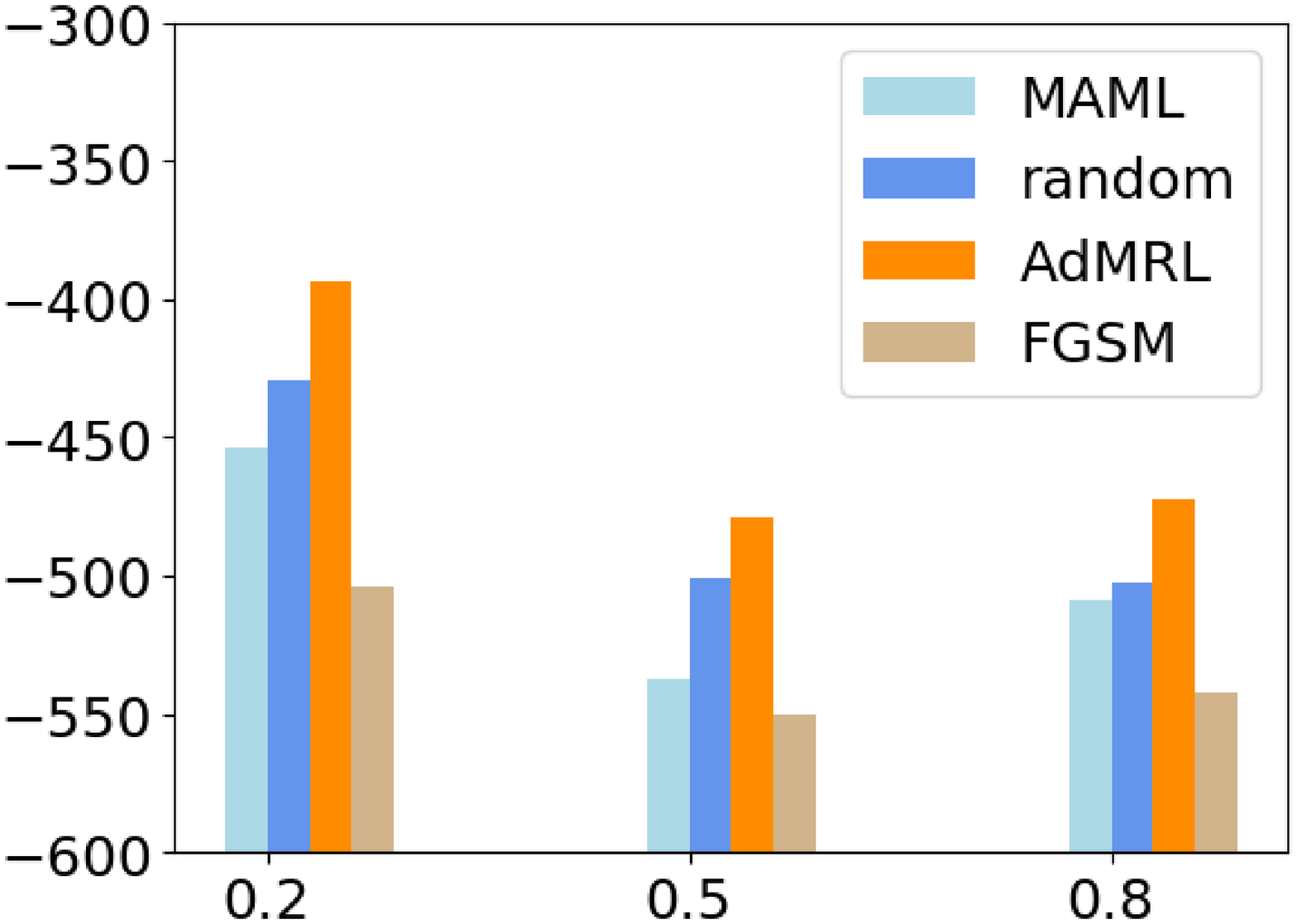}
  \end{minipage}
 }%
 \centering
 \caption{Experiment with adversarial attack. Comparison of different adversarial training methods on MRL on five different environments. It can be observed that our adMRL has better and more stable performance than all other methods. Here, the x-axis shows the strength of attack, we conduct the attack range of 0.2, 0.5 and 0.8 respectively, and the y-axis shows the average total return of the meta-test tasks.}
 \label{fig:2}
\end{figure*}

\begin{figure*}[htbp]
 \centering
 \subfigure[2D-Navigation]{
  \begin{minipage}[t]{0.19\linewidth}
   \centering
   \includegraphics[scale=0.17]{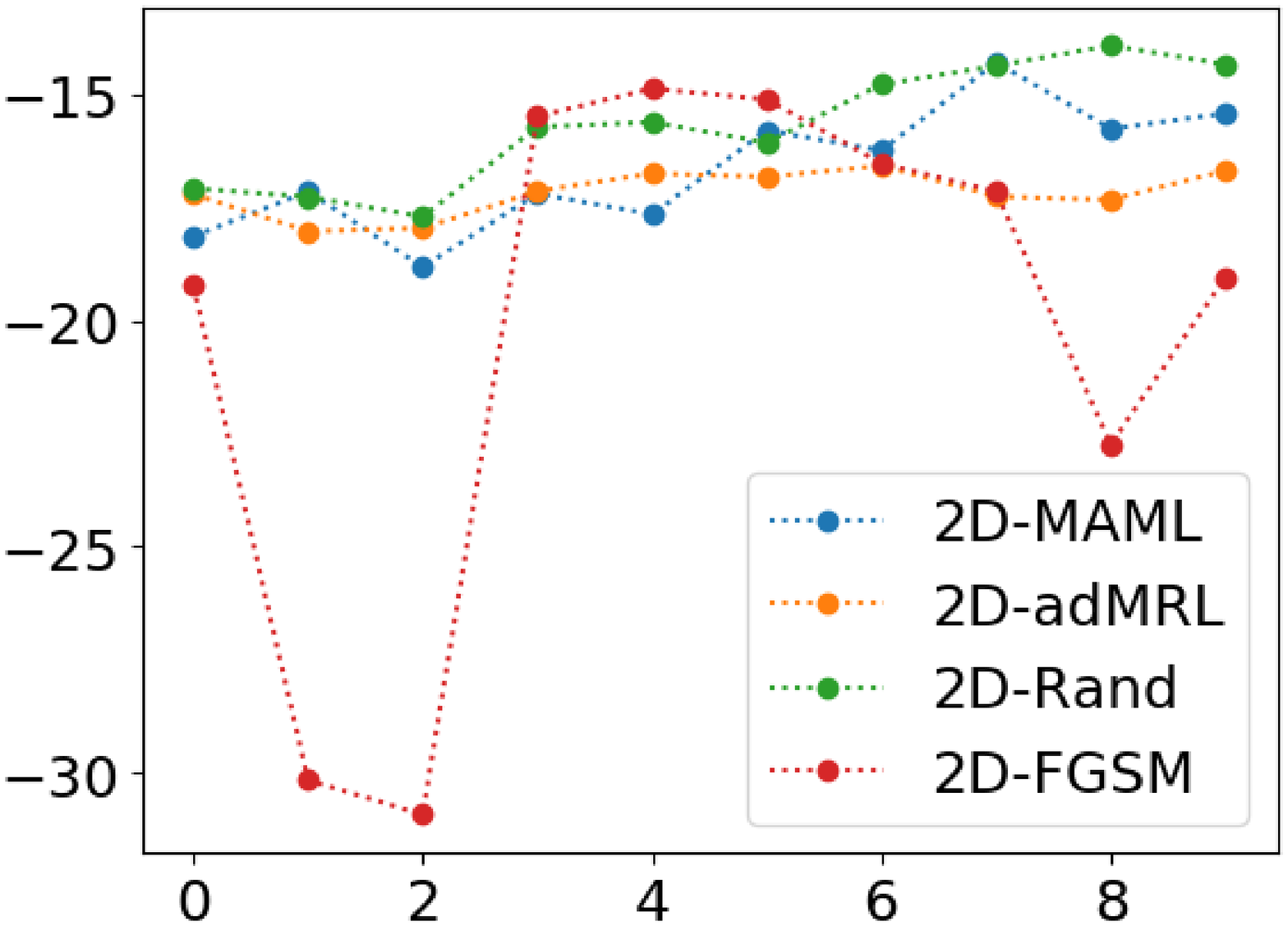}
  \end{minipage}%
 }%
 \subfigure[HalfCheetah-Vel]{
  \begin{minipage}[t]{0.19\linewidth}
   \centering
   \includegraphics[scale=0.17]{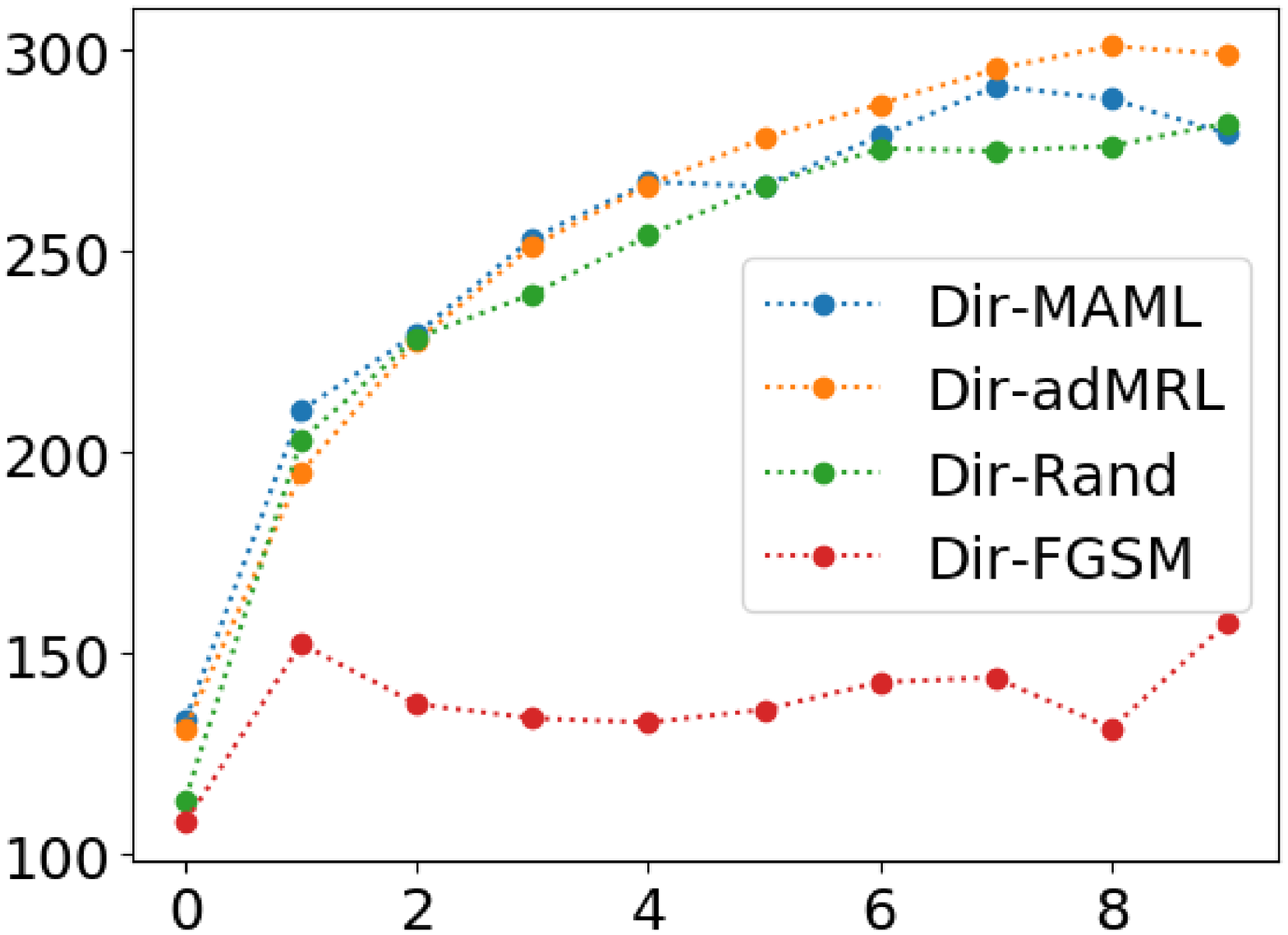}
  \end{minipage}%
 }%
 \subfigure[HalfCheetah-Dir]{
  \begin{minipage}[t]{0.19\linewidth}
   \centering
   \includegraphics[scale=0.17]{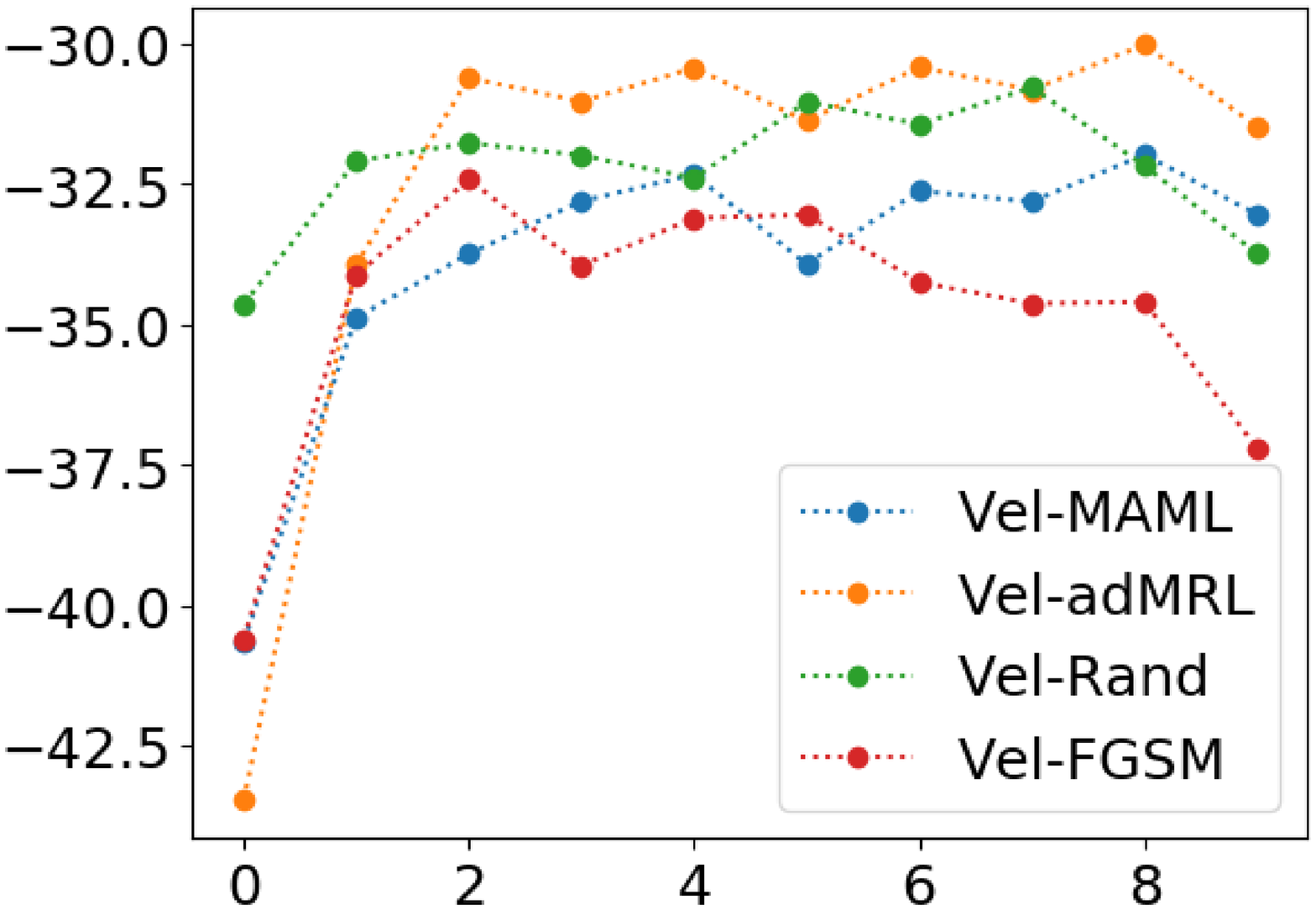}
  \end{minipage}
 }%
 \subfigure[Reacher]{
  \begin{minipage}[t]{0.19\linewidth}
   \centering
   \includegraphics[scale=0.17]{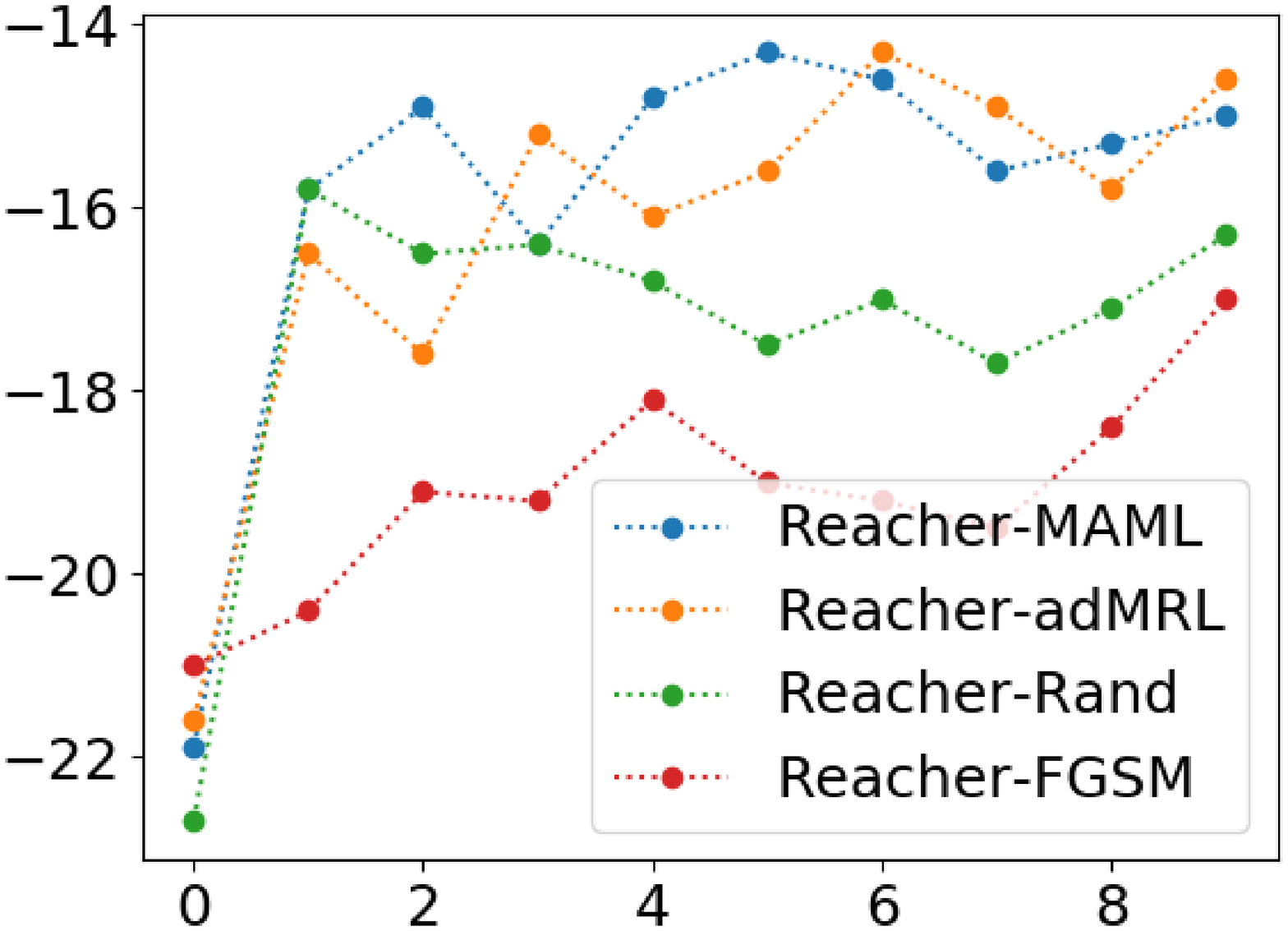}
  \end{minipage}
 }%
 \subfigure[Ant]{
  \begin{minipage}[t]{0.19\linewidth}
   \centering
   \includegraphics[scale=0.17]{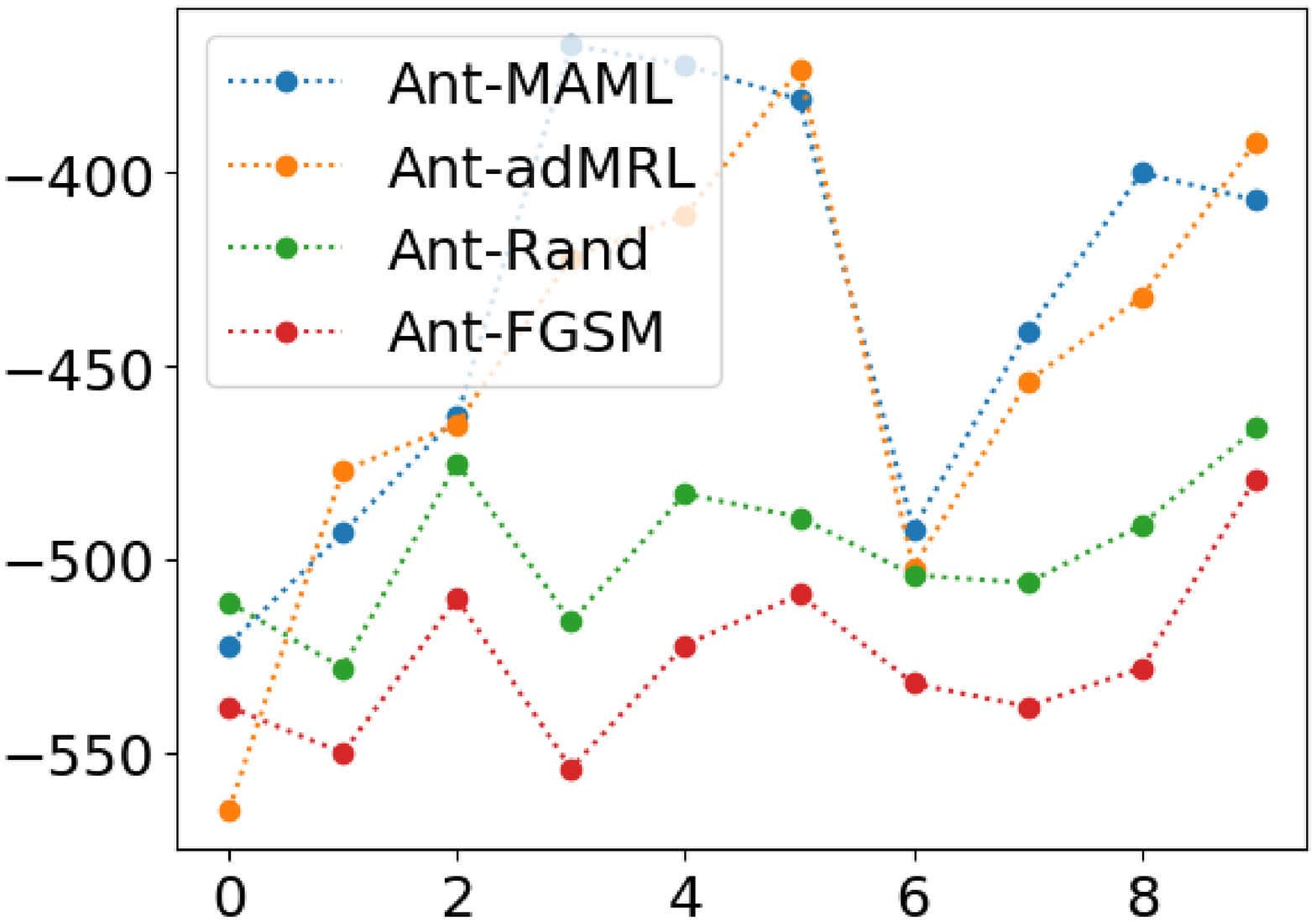}
  \end{minipage}
 }%
 
 \centering
 \caption{Comparison of the convergence curve of different robust training methods on MRL on five different environments under the adversarial attack, the x-axis shows the epoch numbers, y-axis means the average return of the meta-test tasks.}
 \label{fig:3}
\end{figure*}

 We evaluate our adversarial MRL on five continuous control environments of robotic locomotion, simulated via MuJoCo simulator \cite{b19}. These locomotion tasks require adaptation on the reward functions (goal location for 2D-Navigation, walking direction for HalfCheetah-Dir and Reacher, target velocity for HalfCheetah-Vel and Ant). These meta-RL benchmarks were introduced by Finn \cite{b4} and Rothfuss \cite{b15}. We compare our proposed methods to existing policy gradient based MRL algorithm MAML \cite{b4}.

\subsection{Baselines and our model}
 During experiments, we compare four approaches as follows. Hyper parameters of all the baselines are chosen to be optimal.
\begin{itemize}
\item \textbf{MAML:} Model Agnostic Meta Reinforcement Learning proposed by Finn et al \cite{b4}.
\item \textbf{Random Noise training:} In this method, we train the algorithm by adding random Gaussian noise to the observations at the meta-training phase. The "fake" states are generated by adding noise followed normal distribution to states in each time steps. 
     \begin{equation}
     \begin{aligned}
\widetilde{x}=x+\Delta x, \\\Delta x \sim N\left(\mu, \sigma^{2}\right).
\end{aligned}
\end{equation}

\begin{algorithm}
    \caption{Random Noise Training}
    \label{alg2}
 \begin{algorithmic}[1]
  \State Require: $p(T)=$ distribution over tasks. $\alpha,\beta$: step size hyperparameters.
        \State random initialize network parameter $\phi$
  \While {not done do}
  \State Sample batch of tasks $T_{i}\sim p(T)$
        \For{all $T_{i}$}
        \State Sample K trajectories $D=\left \{ \left ( \tilde{x_{1}},a_{1},\cdot \cdot \cdot , \tilde{x_{H}}\right ) \right \}$ using $ f_{\theta }$  and $\phi$ in $T_{i}$. (adding random Gaussian noise to the current status to generate the polluted states)
        \State Evaluate $\triangledown _{\theta}L_{T_{i}}(f_{\theta })$ using D and $ L_{T_{i}}$ in Eq.(1).
        \State Compute adapted parameters with gradient descent use Eq.(3).
        \State Sample K trajectories ${D}'=\left \{ \left ( \tilde{x_{1}},a_{1},\cdot \cdot \cdot , \tilde{x_{H}}\right ) \right \}$ using $ f_{{\theta }'}$  in $T_{i}$. Meanwhile, the  perturbed state $\tilde{x_{i}}$ is also generated by adding the random noise , and the normal states are the input.
        \EndFor
        \State Update $\theta \leftarrow \theta-\beta \bigtriangledown _{\theta }\sum _{T_{i}\sim p(T) }L_{T_{i}}(f_{\theta })$ using each ${D_{i}}'$ and $L_{T^{i}}$
        \State Update $\phi$ using Eq.(5).
        \EndWhile

 \end{algorithmic}
\end{algorithm}

     Here, $\Delta x$ is the generated noise, $x$ is the normal state in current time step, and $\widetilde{x}$ is the perturbed state. In our experiments, here we set the parameter for to standard normal distribution, so $\mu=0$, $\sigma^{2}=1$. We should notice that in practice, we firstly train the agent using "normal" states and parameters until the agent reaches a steady state, then we add noise samples to fine-tune the agent to robustness against model uncertainties. This is to prevent the algorithms to obtain diverge results. In this specific setting, we train the meta agent for 500 steps in total, and we only add the noise after 300 training steps. The corresponding pseudo code is included in the Algorithm 2.
\item \textbf{FGSM training:} In this method, we use FGSM to generate adversarial samples for robust training of MRL \cite{b6}, which is the most classical and influential gradient-based method to generate adversarial samples for adversarial training. The corresponding  pseudo code is shown in Algorithm \ref{alg2}. The objective function of FGSM is given by:
\begin{equation}
\bigtriangleup x=\varepsilon sign(\bigtriangledown _{\theta }J(\theta ,x,y)).
\end{equation}
where $\varepsilon$ controls the amplitude of the perturbation. In our reinforcement learning scenario, the loss function is different from a  supervised learning case, and the perturbed states could be indicated as follows:
\begin{equation}
\bigtriangleup x=\varepsilon sign(\bigtriangledown _{\theta } L_{T_{i}}(f_{\varphi })).
\end{equation}
Similar to other methods, due to the instability characteristic of reinforcement learning, we also use adversarial training only in the fine-tune phase.
\item \textbf{adMRL:} Our proposed adversarial Meta Reinforcement Learner. The hyper parameter settings are as follows: the $\alpha$ is 0.8, the $\beta$ is 0.2. The c is 0.2.
\end{itemize}

\begin{algorithm}[htb]
    \caption{FGSM Training}
    \label{alg2}
 \begin{algorithmic}[1]
  \State Require: $p(T)=$ distribution over tasts. $\alpha,\beta$: step size hyperporameters.
        \State random initialize network parameter $\theta$, GAN parameter $\phi$.
  \While {not done do}
  \State Sample batch of tasks $T_{i}\sim p(T)$
        \For{all $T_{i}$}
        \State Using the current status and  generator of GAN to generate the polluted states.
        \State Sample K trajectories $D=\left \{ \left ( \tilde{x_{1}},a_{1},\cdot \cdot \cdot , \tilde{x_{H}}\right ) \right \}$ using $ f_{\theta }$  and $\tilde{\theta }$ in $T_{i}$. The perturbed states $\tilde{x_{i}}$ is generated by FGSM, and the input are the normal states.
        \State Evaluate $\triangledown _{\theta}L_{T_{i}}(f_{\theta })$ using D and $ L_{T_{i}}$ in Eq.(1).
        \State Compute adapted parameters with gradient descent use Eq.(3).
        \State Sample K trajectories ${D}'=\left \{ \left ( \tilde{x_{1}},a_{1},\cdot \cdot \cdot , \tilde{x_{H}}\right ) \right \}$ using $ f_{{\theta }'}$  in $T_{i}$. Meanwhile, the  perturbed states $\tilde{x_{i}}$ is generated by FGSM, and the input are the normal states.
        \EndFor
        \State Update $\theta \leftarrow \theta-\beta \bigtriangledown _{\theta }\sum _{T_{i}\sim p(T) }L_{T_{i}}(f_{\theta })$ using each ${D_{i}}'$ and $L_{T^{i}}$
        \State Update $\phi$ using Eq.(5).
        \EndWhile

 \end{algorithmic}
\end{algorithm}

\subsection{Evaluation process}
We propose two evaluation approaches to test our methods. Firstly, we propose a naive method of generating random attack, the idea behind this attack is to generate random noise and add it to current state to expect that these "bad" states will cause an agent to take bad actions and decrease the total rewards. The attacks are generated only during the meta-test phase, we set three noise levels in the scale of 0.2, 0.5 and 0.8, respectively.

Then, we test our methods under adversarial attack scenarios. We use our adGAN to generate bad samples in the meta-test phase, and the results show that the adGAN could attack the agent effectively, the total rewards decrease more severely than that under random attack.

\subsection{Experiment Results}
In this section, we present results that show great improvement in robustness due to our proposed adversarial training algorithm. For robust training, we first train a meta agent with MAML-TRPO, and the agent is then becoming robust by fine-tuning based on the adGAN-generating attack. For evaluation, we test this adversarially trained agent on a wide range of parameters of attack. The results show improvement to robustness over MAML.

\textbf{Normal case:} Our adMRL is proved to be an effective meta-reinforcement learner because it leads to quick-learning from limited trajectories for a new task. adMRL maintains satisfactory rewards in the "clean-clean" cases which are the general condition of meta-learning. For instance, under the 2D-Navigation environment, adMRL gives an average return of -12.50, which is very close to that given by MAML (-12.18). Meanwhile, the random training and the FGSM training deliver worse performance which is -13.37 and -21.15 respectively.

\textbf{Random attack case:} As observed in Figure 2, adMRL outperforms other methods under the random noise attack setting. From this result, we can see that our adGAN training has stable and satisfactory performance for almost all cases, which is always better than our baseline MAML-TRPO. An interesting point is that the results of FGSM training and random noise training differ a lot in different environments and  have a quite wide range of changes in different attack. A possible explanation is that these two methods have a relatively large randomness. Random Noise training uses random Gaussian noise to generate training noise, therefore the erratic performance is quite understandable. And for the FGSM training, from the FGSM equation (Equation 15) we can know that the attack are in a fixed scale for every dimension of state. Thus the attack is so strong that the agents fail to learn useful information in some cases. While due to the self-adaptive characteristics of our adGAN training, our method could always obtain gratifying results.
\begin{figure*}[h]
 \centering
 \subfigure[2D-Navigation]{
  \begin{minipage}[t]{0.19\linewidth}
   \centering
   \includegraphics[scale=0.17]{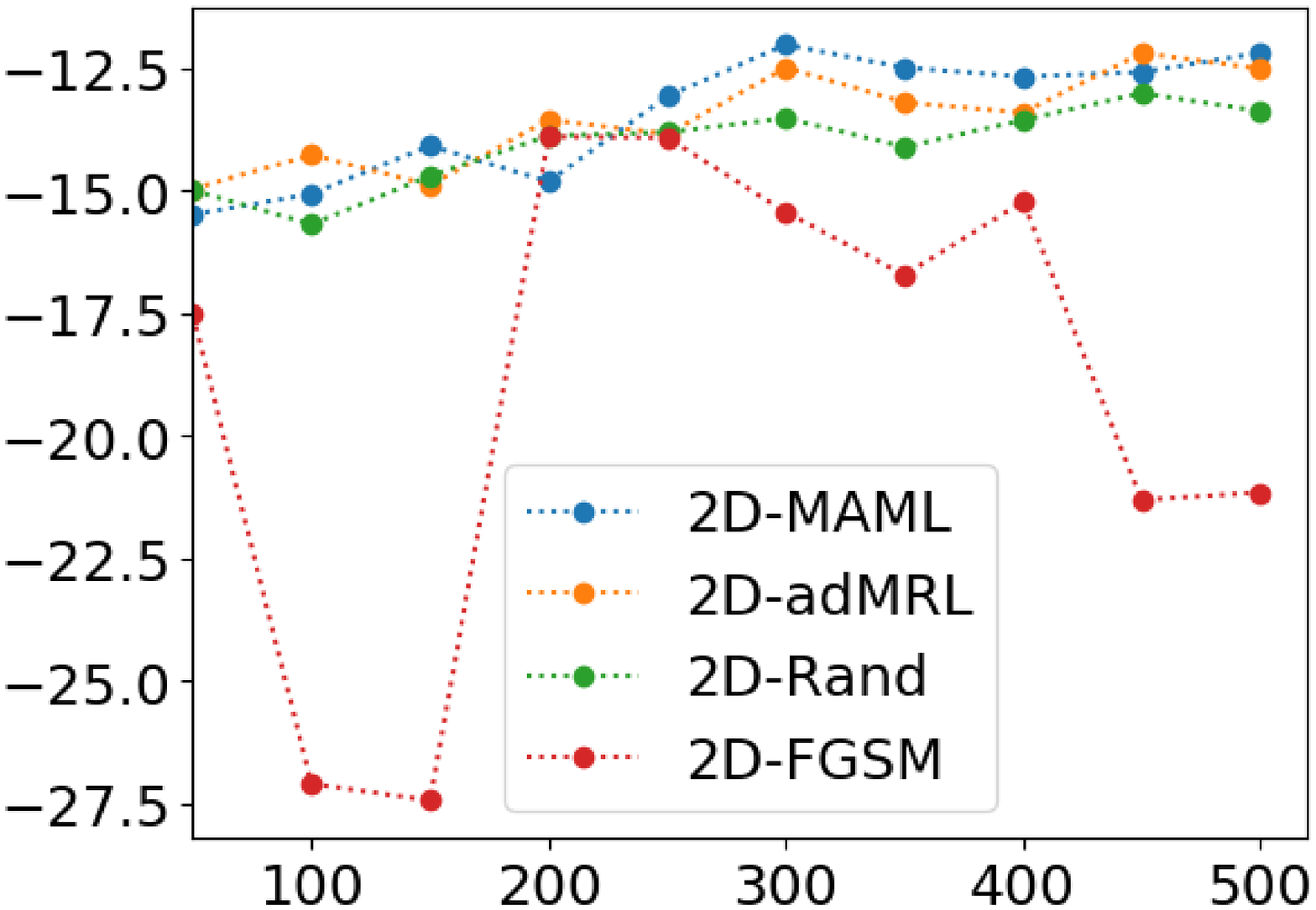}
  \end{minipage}%
 }%
 \subfigure[HalfCheetah-Vel]{
  \begin{minipage}[t]{0.19\linewidth}
   \centering
   \includegraphics[scale=0.17]{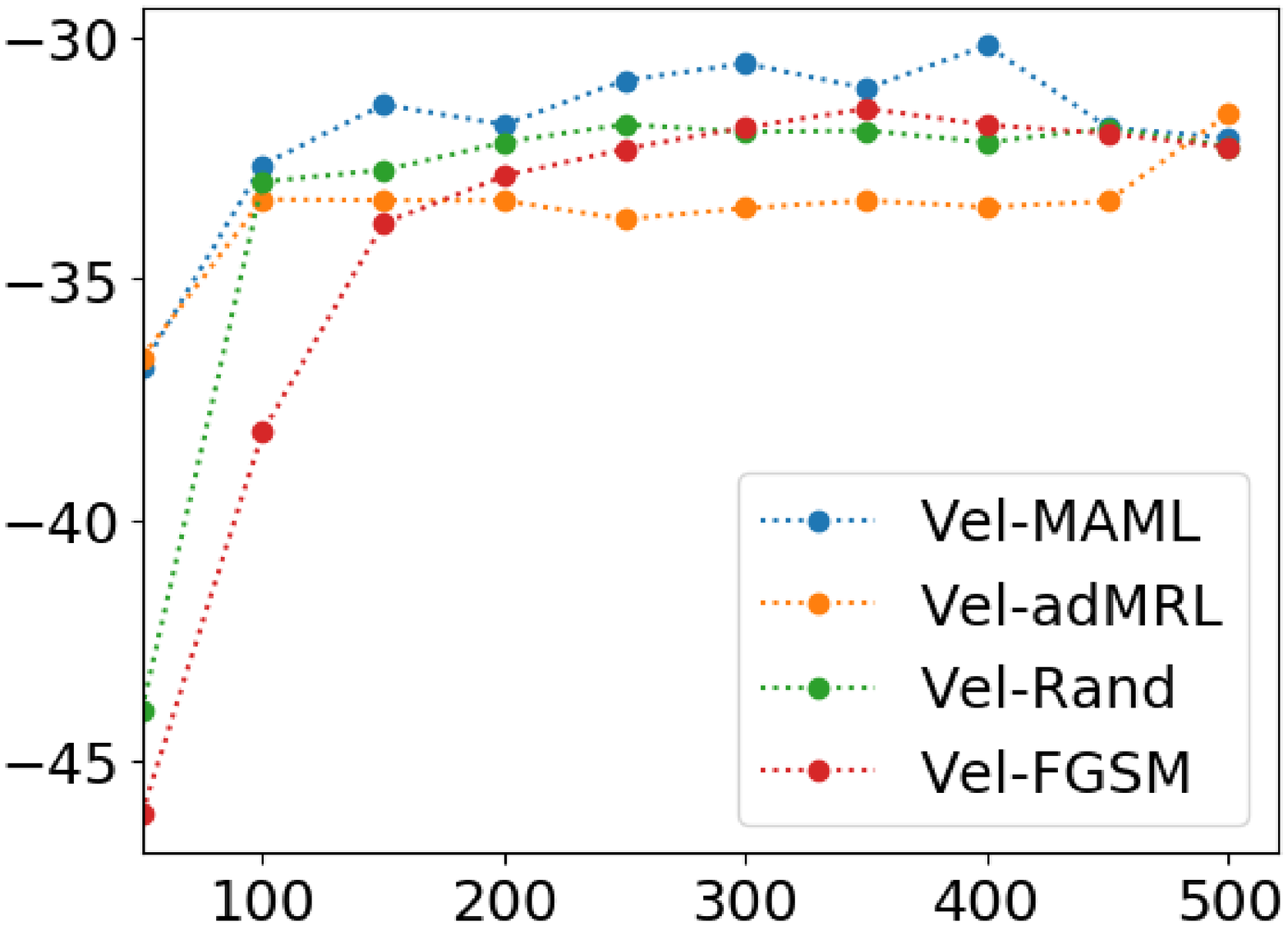}
  \end{minipage}%
 }%
 \subfigure[HalfCheetah-Dir]{
  \begin{minipage}[t]{0.19\linewidth}
   \centering
   \includegraphics[scale=0.17]{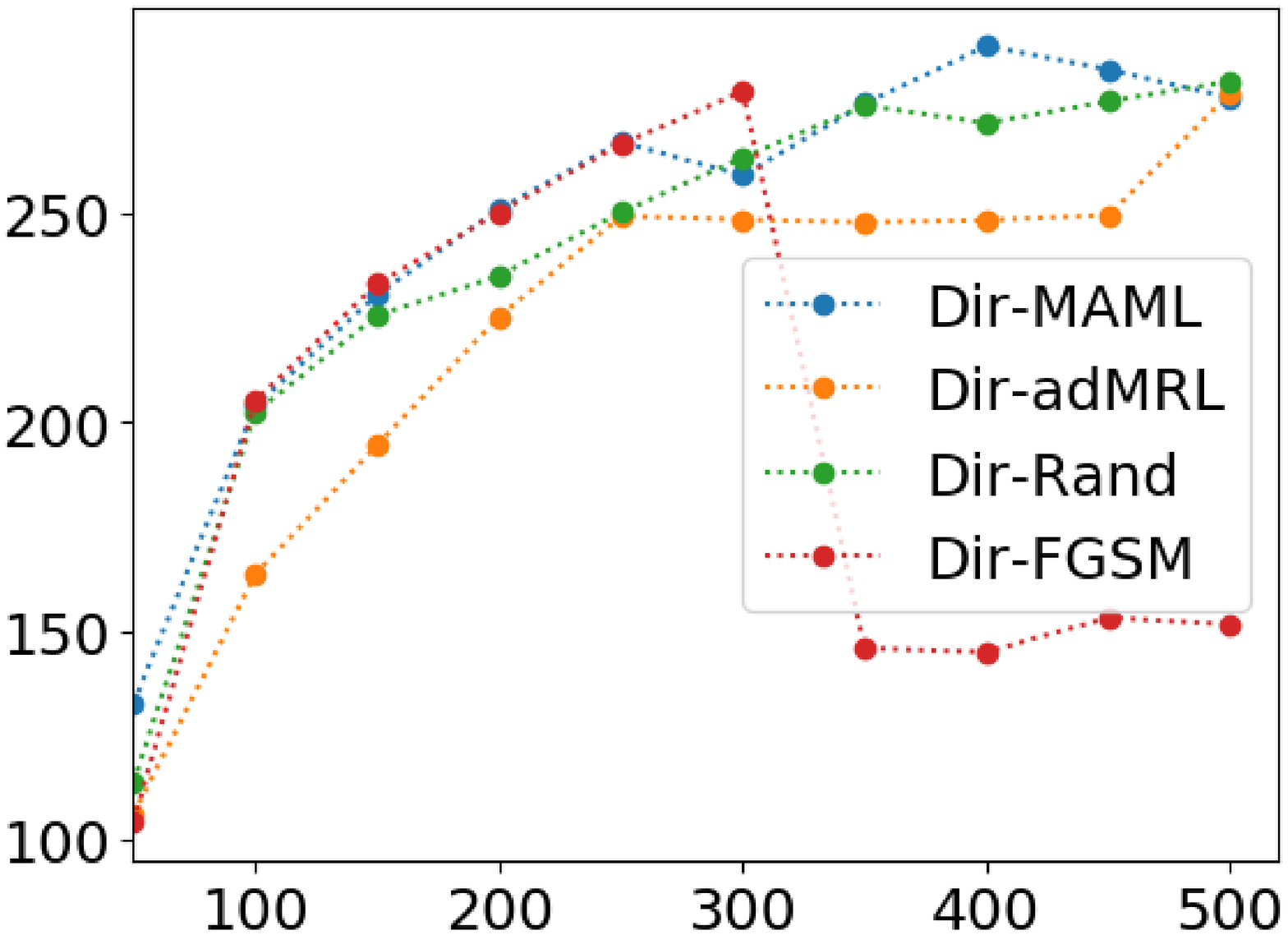}
  \end{minipage}
 }%
 \subfigure[Reacher]{
  \begin{minipage}[t]{0.19\linewidth}
   \centering
   \includegraphics[scale=0.17]{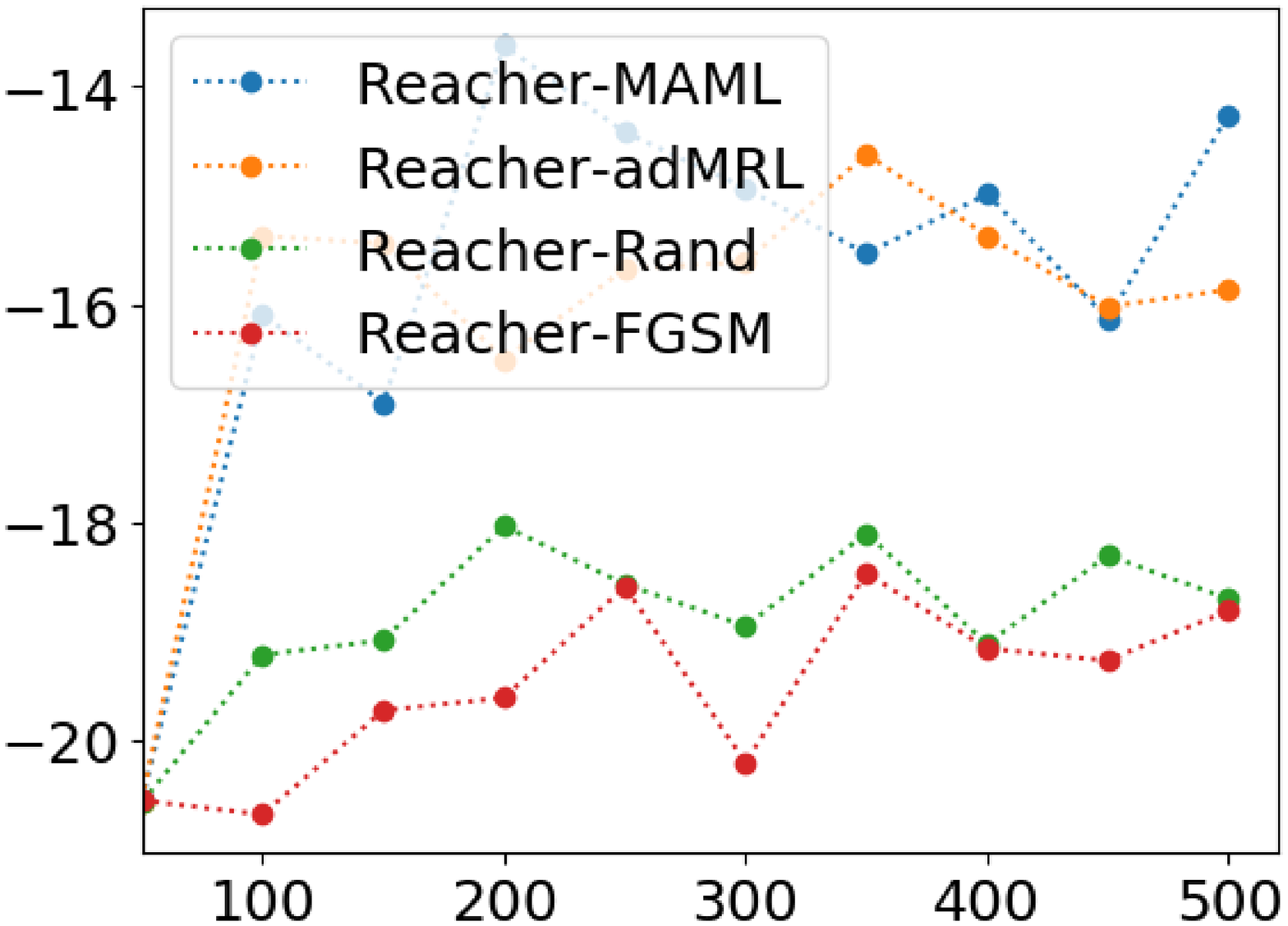}
  \end{minipage}
 }%
 \subfigure[Ant]{
  \begin{minipage}[t]{0.19\linewidth}
   \centering
   \includegraphics[scale=0.17]{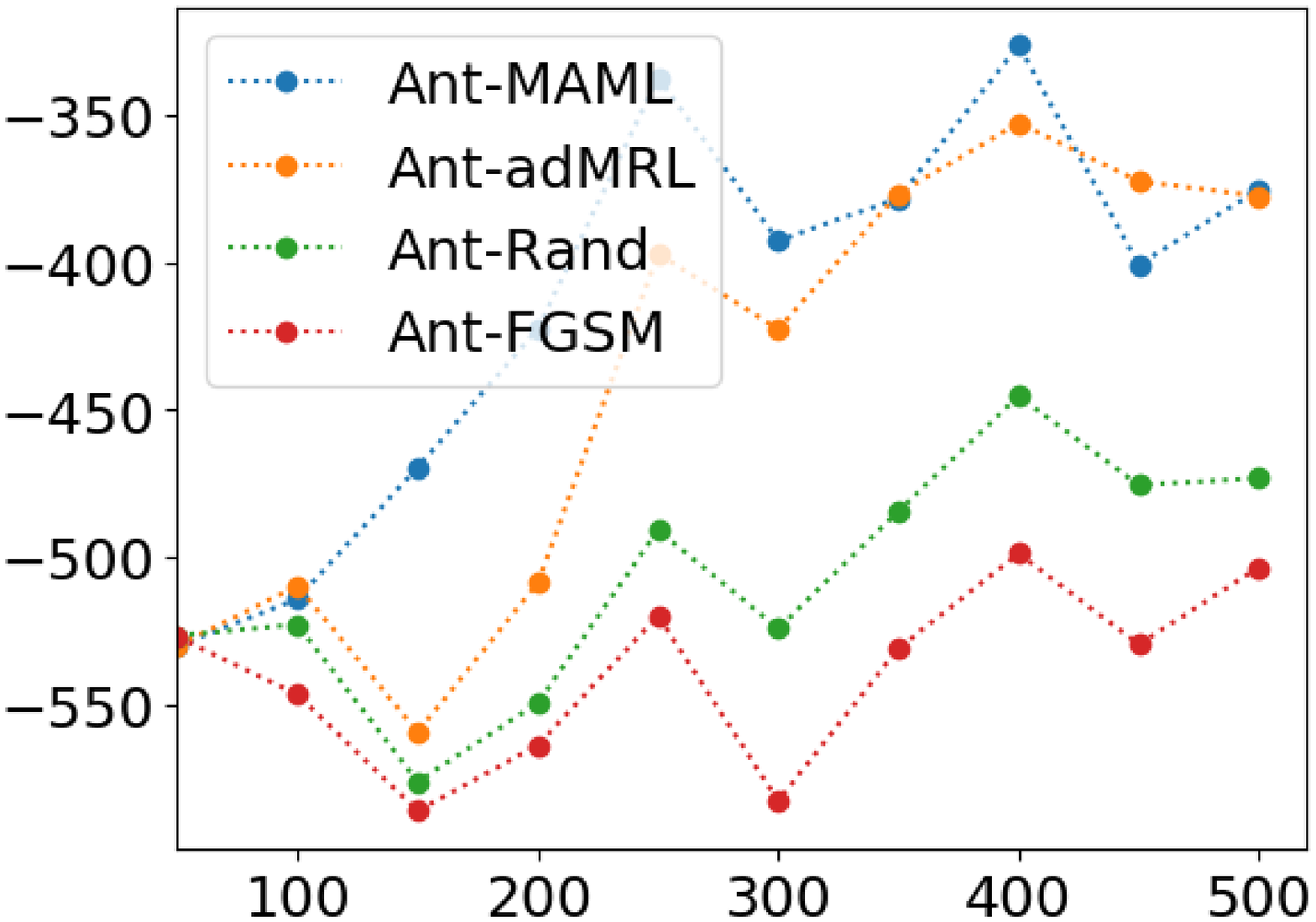}
  \end{minipage}
 }%
 \centering
 \caption{comparison of the convergence curve in normal cases where there are only clean samples, the x-axis shows the epoch numbers, y-axis means the average return (by calculating the mean value of total return). }
 \label{fig:4}
\end{figure*}

\textbf{Adversarial attack case:} adMRL consistently outperforms all other meta-learning methods with adversarial attack in all evaluation environments. For instance, in the HalfCheetah-Dir environment, when injecting adversarial samples in meta-test phase, adMRL achieves total reward of 53.4, while MAML only achieves 50.3, random noise training gives 24.25 and FGSM training only got 0.03. Similar results can also found in other environments. This clearly shows the superiority of the adversarial meta-training procedure of the adMRL.
\begin{table}[t]
\caption{MRL attacked by adversarial attack generated by adGAN}\smallskip\label{symbols}
\centering
 \resizebox{85mm}{12mm}{
{\begin{tabular}[l]{@{}lccccc}
  
\toprule
  Envs & 2D-Navigation & HalfCheetah-Dir & HalfCheetah-Vel  & Reacher & Ant \\
\midrule
  Baseline &-50.2  & 50.3  & -30.8& -561 & -18.7 \\
  GAN & {\bfseries -47.7 } & {\bfseries 53.4 } & {\bfseries -30.2} & {\bfseries -483 } & {\bfseries -17.8}  \\
  Random & -47.8 & 24.3 & -30.9  & -516 & -18.4  \\
  FGSM & -54.1 & 0.03 & -31.4  & -572 & -22.1 \\
\bottomrule
\end{tabular}}
}

\end{table}

\textbf{Stability Analysis:} The adMRL shows quite stable performance compared with the other generalization training approaches. From our experiment results, we can see the random training and FGSM training may gain surprisingly low rewards in certain environments. For example, the random training only achieves 24 of total reward in HalfCheetah-Dir environment, which is even much lower than the baseline (MAML). On the contrary, our adMRL is given strong generalization capacity by the adGAN which accumulates experience during meta-training process and then transfers the learned knowledge to the unseen new tasks in meta-test set. Instead, the random training and FGSM training do not have such capability. Another reason is the instability of reinforcement learning, where inappropriate perturbations could cause the crash of robot training in some cases.

\textbf{Convergence Analysis:} The adMRL is an effective meta reinforcement learner since the adMRL could converge quickly in all environments. Figures \ref{fig:3} and \ref{fig:4} show the variability of total return with the increase of epochs, where we set 500 epochs in total and we record the current returns every 50 epochs. And it can be noticed that our adMRL even outperforms the other three methods in some cases such as HalfCheetah-Dir and HalfCheetah-Vel under the random noise, and it also has the best convergence performance in 2D-navigation and Reacher environments under adversarial attack.

\textbf{ Adaptive adversarial training compared with FGSM:} adMRL has better performance than one other adversarial training method (FGSM). Our experiments show that adMRL training outperforms FGSM training from multiple perspectives, including stability, adaptation and efficiency aspects. Especially,  FGSM training shows severe inefficiency problem in both random noise and adversarial attack condition. For instance,  it only gives 0.03 of total return  which is much lower than baseline in HalfCheetah-Dir under adversarial attack, and it also  performs quite poorly in other environments. And when dealing with random noise,  it also shows unsatisfactory performance in almost all cases except under 2D-Navigation environment, where FGSM achieves the most total return than all other methods.  And from this, we can conclude that the FGSM training is unstable and FGSM lacks generalization ability, because the FGSM could only generate rigid perturbations based on current trajectory, and this kind of perturbation cannot transfer to the test set under the MRL scenario.

\section{Conclusion}
In this paper, we propose a novel method calling adMRL (adversarial Meta Reinforcement Learner) for MRL algorithm, which could learn to regularize the meta-policy for generalization by meta-training an agent under the corrupted environment with the perturbed  states.  Meanwhile, we use the meta-learning method to learn a self-adaptive adversarial samples generator. By optimizing a minmax objective function, the adGAN and MRL enhance each other during training and both of them can get satisfactory results. By comprehensive empirical study on performance evaluation on five robotic locomotion benchmarks, we can draw conclusions as follows: 1) MRL can be easily fooled by adversarial attack even using random noise; 2) adMRL is an effective meta-reinforcement learner even in cases where there are only clean samples; 3) adMRL has better performance than other methods in terms of resisting in different attack (including both random naive attack and generated adversarial attack); 4) It unlocks an interesting topic and sheds light on dealing with limited and even contaminated tasks.
Regarding the future works, we will find more effective way to generate adversarial attack for MRL from empirical and theoretical aspects.

\section{Acknowledgments}
The authors gratefully acknowledge funding support from the Westlake University and Bright Dream Joint Institute for Intelligent Robotics.


\begin{thebibliography}{00}
\bibitem{b1} Carlini N, Wagner D. Adversarial examples are not easily detected: Bypassing ten detection methods[C]//Proceedings of the 10th ACM Workshop on Artificial Intelligence and Security. 2017: 3-14.
\bibitem{b2} Doshi-Velez F, Konidaris G. Hidden parameter markov decision processes: A semiparametric regression approach for discovering latent task parametrizations[C]//IJCAI: proceedings of the conference. NIH Public Access, 2016, 2016: 1432.
\bibitem{b3} Duan Y, Schulman J, Chen X, et al. Rl $^ 2$: Fast reinforcement learning via slow reinforcement learning[J]. arXiv preprint arXiv:1611.02779, 2016.
\bibitem{b4} Finn C, Abbeel P, Levine S. Model-Agnostic Meta-Learning for Fast Adaptation of Deep Networks[C]//ICML. 2017.
\bibitem{b5} Goodfellow I J, Shlens J, Szegedy C. Explaining and harnessing adversarial examples[J]. arXiv preprint arXiv:1412.6572, 2014.
\bibitem{b6} Huang S, Papernot N, Goodfellow I, et al. Adversarial attacks on neural network policies[J]. arXiv preprint arXiv:1702.02284, 2017.
\bibitem{b7} Kos J, Song D. Delving into adversarial attacks on deep policies[J]. arXiv preprint arXiv:1705.06452, 2017.
\bibitem{b8} Lee H B, Nam T, Yang E, et al. Meta Dropout: Learning to Perturb Features for Generalization[J]. arXiv preprint arXiv:1905.12914, 2019.
\bibitem{b9} Lin Y C, Hong Z W, Liao Y H, et al. Tactics of adversarial attack on deep reinforcement learning agents[J]. arXiv preprint arXiv:1703.06748, 2017.
\bibitem{b10} Mendonca R, Geng X, Finn C, et al. Meta-reinforcement learning robust to distributional shift via model identification and experience relabeling[J]. arXiv preprint arXiv:2006.07178, 2020.
\bibitem{b11} Mishra N, Rohaninejad M, Chen X, et al. A simple neural attentive meta-learner[J]. arXiv preprint arXiv:1707.03141, 2017.
\bibitem{b12} Nagabandi A, Clavera I, Liu S, et al. Learning to adapt in dynamic, real-world environments through meta-reinforcement learning[J]. arXiv preprint arXiv:1803.11347, 2018.
\bibitem{b13} Pattanaik A, Tang Z, Liu S, et al. Robust deep reinforcement learning with adversarial attacks[J]. arXiv preprint arXiv:1712.03632, 2017.
\bibitem{b14} Puterman M L. Markov decision processes: discrete stochastic dynamic programming[M].2014.
\bibitem{b15} Rothfuss J, Lee D, Clavera I, et al. Promp: Proximal meta-policy search[J]. arXiv preprint arXiv:1810.06784, 2018.
\bibitem{b16} Sæmundsson S, Hofmann K, Deisenroth M P. Meta reinforcement learning with latent variable gaussian processes[J]. arXiv preprint arXiv:1803.07551, 2018.
\bibitem{b17} Schulman J, Levine S, Abbeel P, et al. Trust region policy optimization[C]//International conference on machine learning. 2015: 1889-1897.
\bibitem{b18} Sutton R S, Barto A G. Reinforcement learning: An introduction[M]. MIT press, 2018.
\bibitem{b19} Todorov E, Erez T, Tassa Y. Mujoco: A physics engine for model-based control[C]//2012 IEEE/RSJ International Conference on Intelligent Robots and Systems. IEEE, 2012: 5026-5033.
\bibitem{b20} Wang J X, Kurth-Nelson Z, Tirumala D, et al. Learning to reinforcement learn[J]. arXiv preprint arXiv:1611.05763, 2016.
\bibitem{b21} Williams R J. Simple statistical gradient-following algorithms for connectionist reinforcement learning[J]. Machine learning, 1992, 8(3-4): 229-256.
\bibitem{b22} Yin C, Tang J, Xu Z, et al. Adversarial meta-learning[J]. arXiv preprint arXiv:1806.03316, 2018.
\bibitem{b23} Xiao C, Li B, Zhu J Y, et al. Generating Adversarial Examples with Adversarial Networks[J].
\bibitem{b24} Rakhsha A, Radanovic G, Devidze R, et al. Policy teaching via environment poisoning: Training-time adversarial attacks against reinforcement learning[C]//International Conference on Machine Learning. PMLR, 2020: 7974-7984.
\bibitem{b25} Goldblum M, Fowl L, Goldstein T. Robust few-shot learning with adversarially queried meta-learners[J]. arXiv preprint arXiv:1910.00982, 2019.
\bibitem{b26} Yan L, Liu D, Song Y, et al. Multimodal Aggregation Approach for Memory Vision-Voice Indoor Navigation with Meta-Learning[J]. arXiv preprint arXiv:2009.00402, 2020.
\bibitem{b27} Liu D, Cui Y, Cao Z, et al. Indoor navigation for mobile agents: A multimodal vision fusion model[C]//2020 International Joint Conference on Neural Networks (IJCNN). IEEE, 2020: 1-8.
\bibitem{b28} Chen Z, Wang D. Multi-Initialization Meta-learning With Domain Adaptation[C]//2021 International Conference on Acoustics, Speech and Signal Processing, {ICASSP} 2021.


\end{thebibliography}
\end{document}